\documentclass[journal]{IEEEtran}
\usepackage{epsfig}
\usepackage{graphicx}
\usepackage{color,xcolor}
\usepackage{amssymb}
\usepackage{booktabs}
\usepackage{multirow}

\ifCLASSINFOpdf
\else
\fi
\usepackage{amsmath}
\begin{document}
%
% paper title
% Titles are generally capitalized except for words such as a, an, and, as,
% at, but, by, for, in, nor, of, on, or, the, to and up, which are usually
% not capitalized unless they are the first or last word of the title.
% Linebreaks \\ can be used within to get better formatting as desired.
% Do not put math or special symbols in the title.
\title{Making Person Search Enjoy the Merits of Person Re-identification}
%
%
% author names and IEEE memberships
% note positions of commas and nonbreaking spaces ( ~ ) LaTeX will not break
% a structure at a ~ so this keeps an author's name from being broken across
% two lines.
% use \thanks{} to gain access to the first footnote area
% a separate \thanks must be used for each paragraph as LaTeX2e's \thanks
% was not built to handle multiple paragraphs
%

\author{Chuang Liu,
        Hua Yang,
        Qin Zhou,
        and~Shibao Zheng% <-this % stops a space
\thanks{Chuang Liu, Hua Yang, and Shibao Zheng are with the Institute of Image Communication and Network Engineering, Department of Electronic Engineering, Shanghai Jiao Tong University, Shanghai 200240, China and Shanghai Key laboratory of Digital Media Processing and Transmission, Shanghai Jiao Tong University, China (Corresponding authors: Hua Yang and Shibao Zheng).}% <-this % stops a space
\thanks{Qin Zhou are with the Institute of Medical Robotics, Shanghai Jiao Tong University, Shanghai 200240, China.}% <-this % stops a space
}
\maketitle

% As a general rule, do not put math, special symbols or citations
% in the abstract or keywords.
\begin{abstract}
Person search is an extended task of person re-identification (Re-ID). However, most existing one-step person search works have not studied how to employ existing advanced Re-ID models to boost the one-step person search performance due to the integration of person detection and Re-ID. To address this issue, we propose a faster and stronger one-step person search framework, the Teacher-guided Disentangling Networks (TDN), to make the one-step person search enjoy the merits of the existing Re-ID researches. The proposed TDN can significantly boost the person search performance by transferring the advanced person Re-ID knowledge to the person search model. In the proposed TDN, for better knowledge transfer from the Re-ID teacher model to the one-step person search model, we design a strong one-step person search base framework by partially disentangling the two subtasks. Besides, we propose a Knowledge Transfer Bridge module to bridge the scale gap caused by different input formats between the Re-ID model and one-step person search model. During testing, we further propose the Ranking with Context Persons strategy to exploit the context information in panoramic images for better retrieval. Experiments on two public person search datasets demonstrate the favorable performance of the proposed method.
\end{abstract}

% Note that keywords are not normally used for peerreview papers.

% For peer review papers, you can put extra information on the cover
% page as needed:
% \ifCLASSOPTIONpeerreview
% \begin{center} \bfseries EDICS Category: 3-BBND \end{center}
% \fi
%
% For peerreview papers, this IEEEtran command inserts a page break and
% creates the second title. It will be ignored for other modes.
\IEEEpeerreviewmaketitle

\section{Introduction}
% The very first letter is a 2 line initial drop letter followed
% by the rest of the first word in caps.
% 
% form to use if the first word consists of a single letter:
% \IEEEPARstart{A}{demo} file is ....
% 
% form to use if you need the single drop letter followed by
% normal text (unknown if ever used by the IEEE):
% \IEEEPARstart{A}{}demo file is ....
% 
% Some journals put the first two words in caps:
% \IEEEPARstart{T}{his demo} file is ....
% 
% Here we have the typical use of a "T" for an initial drop letter
% and "HIS" in caps to complete the first word.

\begin{figure}[t]
\begin{center}
   \includegraphics[width=1.0\linewidth]{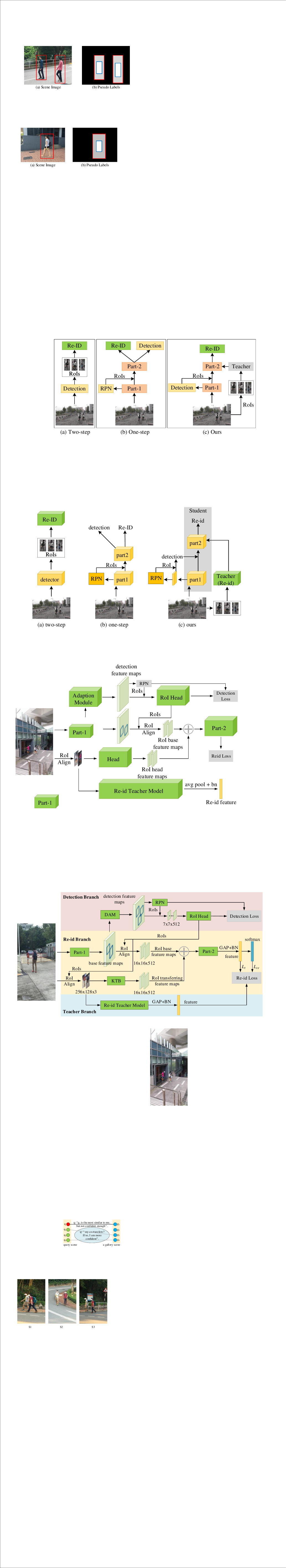}
\end{center}
\caption{Person search frameworks. (a) The two-step framework. It separates person search into two independent subtasks, namely person detection and person Re-ID. (b) The widely-used one-step framework. It is based on the Faster R-CNN framework and shares the whole backbone between the two subtasks. (c) The proposed framework. It partially disentangles the two subtasks to solve the conflict between them and is guided by a Re-ID teacher model to learn more discriminative Re-ID features.}
\label{fig:structure_comparison}
\end{figure}

\IEEEPARstart{P}{erson} search task aims to locate the given query persons in the panoramic images. Different from person re-identification (Re-ID), person search unifies the person detection and re-identification tasks, making it more suitable for real-world applications.
Existing CNN-based person search methods can be divided into two categories: two-step framework and one-step framework. As shown in Fig.~\ref{fig:structure_comparison}, the two-step framework separates person search into two independent subtasks, person detection and person Re-ID, while the one-step framework integrates the detection and Re-ID subtasks into a unified and end-to-end trainable framework with shared networks. In the two-step framework, the Re-ID and detection models are trained separately, which ignores the correlations between them, leading to sub-optimal results. Therefore, we focus on the one-step framework to exploit the correlations between the two subtasks.

As an extended task of person Re-ID, it is intuitive that the one-step person search can benefit from the existing person Re-ID researches. However, due to model structure differences between the one-step person search model and independent Re-ID model, most previous one-step person search works have not studied how to make the advanced Re-ID models benefit the one-step person search model. 
Inspired by the thought of Knowledge Distillation~\cite{hinton2015distilling}, we propose a new one-step person search framework, the Teacher-guided Disentangling Networks (TDN), to make the one-step person search enjoy the merits of the powerful state-of-the-art Re-ID models. 
Specifically, we introduce a teacher branch into the TDN to guide the learning of the one-step person search model. 
The teacher branch is an independent Re-ID model.
To generate discriminative enough Re-ID features as the guidance of the one-step person search model, a powerful state-of-the-art Re-ID model is chosen and pre-trained following classical Re-ID protocols. 

However, there still remain challenges in transferring knowledge from the Re-ID teacher model to the one-step person search model.
Firstly, as pointed out by~\cite{chen2018person}, person detection focuses on the commonness of various persons, while person Re-ID pays attention to the uniqueness among different persons, leading to conflicting optimization targets in the one-step framework depicted in Fig.~\ref{fig:structure_comparison}(b). This conflict prevents the one-step person search model from learning discriminative Re-ID features.
Secondly, different from the traditional KD pipeline where the student model and teacher models have the same input, the Re-ID teacher model takes fixed-scale person patches as input, while the one-step person search model takes the panoramic images as input, which leads to huge scale variations. As analyzed in~\cite{goodfellow2016deep}, the CNN model is not robust against scale variations and can not learn scale-invariant features, making it difficult to directly transferring the strong prior knowledge learned on fixed-scale inputs to the one-step person search model. 

To tackle these issues for effective knowledge transfer, firstly, we design a new strong one-step person search base framework for the proposed TDN, as shown in Fig.~\ref{fig:structure_comparison}(c). The proposed new base framework partially disentangles the two subtasks by only sharing part convolutional networks between them. In this way, the conflict of optimization targets between two subtasks can be reduced to allow the one-step person search model to learn discriminative Re-ID features from the teacher branch more easily. 
Secondly, we propose a Knowledge Transfer Bridge (KTB) module to solve the scale variations for knowledge transfer. The KTB processes the same inputs as the teacher branch, and fuses its output feature maps into the one-step person search. The KTB functions as a bridge to help transfer knowledge from the teacher model to the one-step person search model. Altogether, the proposed TDN can effectively learn from the powerful Re-ID teacher model to generate discriminative Re-ID features. As a result, the proposed TDN demonstrates faster and stronger than the previous one-step framework. 

Besides the proposed TDN, we also improve the ranking process. The widely-used ranking process only considers the individual similarity between persons, ignoring the context persons provided by the panoramic scene images.
To exploit the context information in the panoramic scene images in the ranking phase, we design a Ranking with Context Persons (RCP) strategy which takes context persons as additional cues and generates better ranking results. Specifically, in the proposed RCP, the co-occurrence index score is proposed to measure the correlation between the target person and a context person. The total co-occurrence index scores are taken as a supplement to the individual similarity score to generate the final similarity score for the ranking process.

To summarize, the contributions of this paper are as follows: 
\begin{itemize}
    \item We propose the Teach-guided Disentangling Networks (TDN) to make the one-step person search model enjoy the merits of the powerful state-of-the-art Re-ID models. The proposed TDN can significantly boost the person search performance by integrating the knowledge of powerful Re-ID models;
    \item We design the Ranking with Context Person (RCP) strategy to exploit the context persons for better retrieval performance.
    \item  Experiments on two person search benchmarks demonstrate that the proposed method achieves much higher performance compared with state-of-the-art methods. 
\end{itemize}

\section{Related Work}

\textbf{Person Re-ID.} Person Re-ID has been studied for many years and achieved great progress. Some works~\cite{bak2017one,liu2017stepwise,hermans2017defense,chen2017beyond,yu2019reference,chang2019distribution,yan2021beyond} focus on designing effective metric principles for person Re-ID. Some works~\cite{chen2018group,chen2018video,song2018region,chang2020weighted,luo2019strong,chen2020salience,zhang2020relation,gong2021lag} pay attention to developing novel CNN models to learn discriminative features. 
As an extended task of person Re-ID, person search task should benefit from the existing person Re-ID researches.

\textbf{Person Search.} In recent years, person search task has already received a lot of attention due to the integration of person detection and Re-ID. Many two-step structure methods are proposed in \cite{zheng2017person,lan2018person,chen2018person,han2019re,wang2020tcts}.
Zheng et al.~\cite{zheng2017person} first propose the two-step person search framework and evaluate the impact of combinations of various person detector and person Re-ID models. 
Lan et al.~\cite{lan2018person} propose the Cross-Level Semantic Alignment to solve the multi-scale challenge by combining cross-level feature maps with the Faster R-CNN~\cite{ren2015faster} as person detector. 
Wang et al.~\cite{wang2020tcts} propose a Task-Consist
Two-Stage person search framework including an
identity-guided query detector to generate query-like person detections and a Detection Results Adapted Re-ID model to make the Re-ID model adapted to the detections.

Different from the above two-step methods, some researchers propose to tackle person search using the one-step structure methods~\cite{xiao2017joint,li2018spatial,munjal2019query,hong2019cascaded,munjal2019query,dong2020bi,chen2020norm}. 
Xiao et al.~\cite{xiao2017joint} first propose the one-step framework based on the Faster R-CNN and design the OIM loss function to solve the ill-conditioned training problem. 
Munjal~et al.~\cite{munjal2019query} propose a query-guided one-step person search framework which employs the query to generate query-relevant proposals.
Chen et al.~\cite{chen2020norm} propose the Norm-Aware Embedding method which uses the norm and angle of the person embedding to conduct person detection and person re-identification, respectively.

\begin{figure*}[htp]
\begin{center}
   \includegraphics[width=0.95\linewidth]{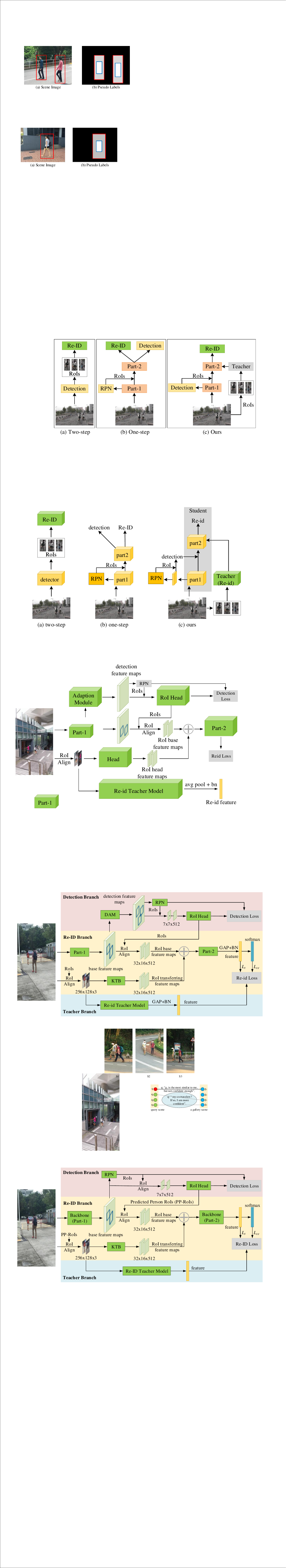}
\end{center}
\caption{Illustration of the proposed TDN. The TDN contains three branches, a detection branch, a Re-ID branch, and a teacher branch. The Re-ID branch is utilized to extract Re-ID features for the person RoIs predicted by the detection branch. The teach branch is employed to supervise the learning of the Re-ID branch and is removed during the reference.}
\label{fig:framework}
\end{figure*}

\textbf{Knowledge Distillation.} 
Hinton et al.~\cite{hinton2015distilling} propose the Knowledge Distillation (KD) to compress complex large models into a small model which can keep comparative performance. 
Romero et al.~\cite{romero2014fitnets} introduce the intermediate representations from the teacher model as hints to improve the distillation training procedure. 
Zagoruyko et al.~\cite{zagoruyko2016paying} propose to transfer knowledge by attention maps rather than feature maps.
Yim et al.~\cite{yim2017gift} define the flow of solution procedure (FSP) matrix to represent the flow between two layers and propose to transfer distilled knowledge by minimizing the distance between the teacher and student FPS matrices. 
Generally, these knowledge distillation works focus on how to effectively transfer knowledge from the teacher model to the student model, where both the teacher and student models solve the same task and have the same input. 

Munjal et al.~\cite{munjal2019knowledge} also propose to apply knowledge distillation to person search. However, their method distills the knowledge of a pre-trained person detector to the one-step OIM person search model to improve the person detection. Differently, in this paper, inspired by the thought of knowledge distillation, we propose a new one-step person search framework (the TDN) to integrate the state-of-the-art person Re-ID models to promote the development of one-step person search. Different from the above-mentioned knowledge distillation works, the proposed TDN framework aims to transfer the knowledge from a single-task teacher model taking cropped person patches as inputs to a multi-task student model taking panoramic images as inputs, which brings some new challenges to the knowledge distillation process. For effective knowledge distillation, we take some measures to overcome the new challenges in the proposed TDN framework.

\section{Method}

In this section, we will elaborate on the proposed TDN framework as well as the RCP ranking strategy. As illustrated in Fig.~\ref{fig:framework}, the TDN framework includes a detection branch, a Re-ID branch, and a teacher branch. To perform person search, the detection branch is used to locate persons in the panoramic images based on the outputs of the Part-1, then Re-ID features are extracted from the Re-ID branch to represent each predicted person. The teacher branch is used to guide the Re-ID branch to learn discriminative features in the training phase. Finally, the RCP ranking method provides the ranking list for each query image.

\subsection{Teacher-guided Disentangling Networks}
In the proposed TDN, to make the one-step person search model learn more discriminative Re-ID features from the powerful Re-ID teacher model, we design a new one-step base framework, namely the Partially Disentangled Framework, and propose a Knowledge Transfer Bridge (KTB) module.
As shown in Fig.~\ref{fig:framework}, the proposed TDN consists of a Re-ID branch, a detection branch and a teacher branch. The Re-ID branch and detection branch form the Partially Disentangled Framework which carries out the person detection and re-identification subtasks. The KTB module is introduced between the Re-ID branch and the teacher branch to help the one-step person search model learn from the teacher model more easily. The details are introduced in the following parts.

\textbf{Partially Disentangled Framework.}
For the Re-ID branch, we adopt the ResNet50-IBN-a~\cite{pan2018two} as the backbone. The ResNet50-IBN-a is divided into two parts, Part-1 and Part-2. The Part-1 is composed of layers from the conv1 to conv3\_x, while the Part-2 includes layers from conv4\_x to conv5\_x. Following the Part-2, a Global Average Pooling (AVG) layer and a Batch Normalization (BN) layer are used to generate the final Re-ID features. Finally, a softmax layer classifies person identities in the training phase.

For the detection branch, we do not adopt the widely-used one-step framework~\cite{xiao2017joint} shown in Fig.~\ref{fig:structure_comparison}(b).
As analyzed in the introduction, the detection and Re-ID subtasks have a conflict of optimization targets in this widely-used one-step framework~\cite{xiao2017joint} which shares all convolutional networks between two subtasks. The conflict prevents the Re-ID subtask from learning discriminative Re-ID features for each detected person RoI.
To address this problem, we propose to partially disentangle the two branches to retain both relevance and independence between them. 
Specifically, in the proposed architecture, only the Part-1 network is shared between the detection and Re-ID branches (as illustrated in Fig.~\ref{fig:framework}), while Part-2 is only kept for the Re-ID branch to learn discriminative features for person RoI detected by the detection branch. 

Based on the feature maps generated by the Part-1, a standard RPN~\cite{ren2015faster} is built to generate possible person RoIs. 
And the RoI Align~\cite{he2017mask} operation is adopted to crop the feature maps of the RoIs into fixed-size feature maps ($7\times7\times512$). 
These 3-D feature maps are then flattened into 1-D feature vectors before fed into the RoI Head. 
The RoI Head is composed of two fully connected (FC) layers with 1024 units. Finally, a classification layer and a coordinate regression layer are employed to predict the probability that the RoIs are persons and refine their box coordinates, respectively. The Non Maximum Suppression (NMS) is applied to the proposals generated from the RoI Head to output the final predicted person RoIs. The Part-2 only extracts Re-ID features for the final predicted person RoIs rather than all the RoIs from the RPN, which improves the running speed.

\textbf{Teacher-guided Learning with KTB.}
For the teacher branch, we choose a state-of-the-art Re-ID model, the strong Re-ID baseline model~\cite{luo2019strong}, as the default teacher model. To pre-train the Re-ID teacher model, we construct a Re-ID style training set by cropping all labeled person patches from the panoramic images and resizing them to $256\times128\times3$ ones with the bilinear interpolation. Then, we pre-train the strong Re-ID teacher model following its training settings~\cite{luo2019strong}.
During the training of the TDN, the pretrained teacher model is utilized to extract features for each predicted person RoI to supervise the learning of the one-step person search model.
Please note that the teacher branch is only used in the TDN training phase, which will not bring an extra computational burden during inference. 

In the one-step person search framework, different from traditional fixed-size Re-ID inputs, features are extracted from predicted person RoIs with arbitrary sizes. 
As analyzed in~\cite{goodfellow2016deep}, the CNN model is not robust against scale variations, making it difficult to directly transfer the strong prior knowledge learned on fixed-size inputs to the Re-ID branch in the one-step person search model. 
To ensure effective knowledge transfer from the strong Re-ID teacher model to the Re-ID branch, we further propose the KTB module for the Re-ID branch to reduce scale variations between the Re-ID teacher model and the Re-ID branch. 
Specifically, denote a predicted person RoI with varying sizes as $x$, we introduce $\hat x$ as a supplementary input of the Re-ID branch, where $\hat x$ is a fixed-size ($256\times128 \times 3$) image interpolated from $x$ by the RoI Align operation. Then, the KTB module processes the supplementary input $\hat x$ to generate the $32\times 16 \times 512$ RoI transferring feature maps. The KTB has the same structure as Part-1 in the Re-ID branch, but does not share parameters with Part-1 to avoid entanglement with the detection task. 

Next, we combine the RoI base feature maps extracted from the Part-1 with the RoI transferring feature maps generated from the KTB module. 
As the RoIs predicted by the detection branch are of variable sizes, to allow fusion with the fixed-size RoI transferring feature maps from the KTB, the RoI base feature maps are interpolated to $32\times 16\times 512$ ones using the RoI Align operation (as shown in Fig.~\ref{fig:framework}). Then, the RoI base feature maps and the RoI transferring feature maps can be fused by pixel-wise addition. The fused feature maps contain feature maps extracted from the same inputs as the teacher model, and consequently can help the Re-ID branch to learn from the teacher model more easily by reducing scale variations.

Finally, in the Re-ID branch, the Part-2 together with the "GAP+BN" layer is utilized to further process the fusion feature maps to generate the final 2048-D Re-ID features, followed by a softmax classifier. Additionally, as shown in Fig.~\ref{fig:framework}, there are two RoI Align operations in the Re-ID branch. One is performed on the base feature maps, and the other is conducted on the input scene image. Please kindly note that these two RoI Align operations use the same person RoIs predicted by the detection branch in both the training and inference stages.

\textbf{Training Loss.}
Following the Faster R-CNN~\cite{ren2015faster}, we employ the RPN training losses ($L_{cls}^{rpn}$ and $L_{reg}^{rpn}$) and RoI Head training losses ($L_{cls}$ and $L_{reg}$) to train the detection branch. The total detection loss $L_{det}$ is defined as:
\begin{equation}
    L_{det}=L_{cls}^{rpn}+L_{reg}^{rpn}+L_{cls}+L_{reg}.
    \label{eq.4}
\end{equation}

For the Re-ID branch, we adopt the following loss function to perform knowledge transfer.
\begin{equation}
    L_t= \frac{1}{K}\sum_{x\in \text{RoIs}} \|f_s\left(x,\hat x\right)-f_t\left(\hat x\right)\|^2,
    \label{eq.1}
\end{equation}
where $K$ is the number of the final predicted person RoIs in a mini-batch, $f_t(\hat{x})$ is the L2-normalized Re-ID supervision feature from the teacher model, and $f_s(x,\hat{x})$ is the L2-normalized Re-ID feature from the Re-ID branch. Besides, we also introduce the cross-entropy loss $L_{ce}$. Overall, the total loss for the Re-ID branch is as follows:
\begin{equation}
    L_{reid} = wL_t+L_{ce},
    \label{eq.2}
\end{equation}
where $w$ is a weight factor to balance between the two components of $L_{reid}$.
In the training phase, to let the teacher model lead the learning of Re-ID branch, we empirically set $w$ as follows:
\begin{equation}
    w=\begin{cases}
    5, & epoch< 15\\
    11-0.4\times epoch, & 15\leq epoch < 25 \\
    1, & \text{otherwise}
    \end{cases}.
\label{eq.3}
\end{equation}
There are labeled and unlabeled person RoIs in person search datasets.
Please kindly note that the $L_t$ loss is applied to both the labeled and unlabeled person RoIs, while the $L_{ce}$ loss is only applied to the labeled person RoIs. 

The total training loss for the proposed TDN is the combination of the detection loss $L_{det}$ and  the Re-ID loss $L_{reid}$, which is mathematically formulated as follows,
\begin{equation}
    L=L_{reid}+L_{det}.
    \label{eq.5}
\end{equation}

\subsection{Ranking with Context Persons}

\begin{figure}[htp]
\begin{center}
   \includegraphics[width=1.0\linewidth]{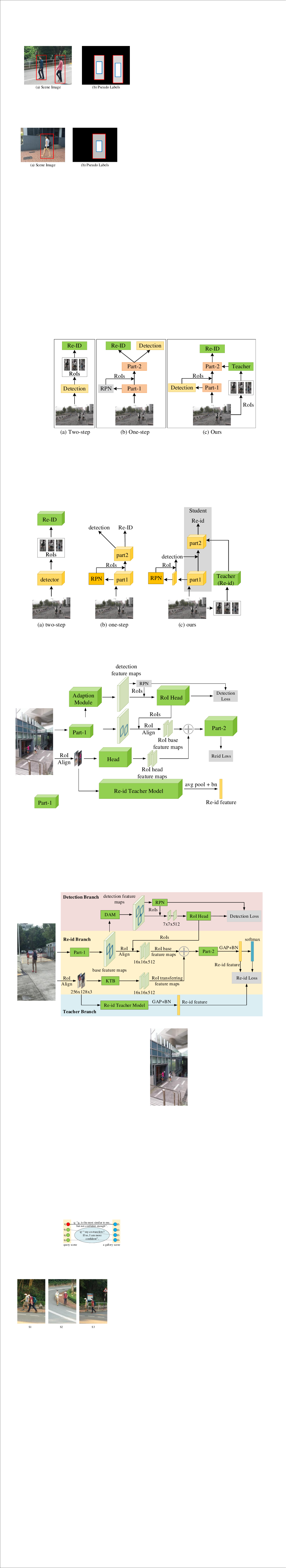}
\end{center}
\caption{Illustration of the fact that context persons tend to simultaneously appear in more than one scenes.}
\label{fig:rcp}
%\vspace{-0.3cm}
\end{figure}

Different from the traditional Re-ID task, the person search task provides the panoramic scene images which contain context persons (co-travelers).
Mazzon et al. \cite{mazzon2013detection} point out that persons are likely to walk in groups, which means that context persons of the query person are likely to simultaneously appear with him in more than one gallery scenes (as shown in Fig.~\ref{fig:rcp}). Thus, the context persons of the query person can be regarded as additional cues to help search for the query. In this paper, we propose an effective ranking method to take advantage of the context cues for better ranking results.

Given a query scene image containing the query person $q$ and context persons $q_1,\cdots,q_N$ and a gallery scene including gallery persons $g_1,g_2,\cdots,g_M$, it is a conventional practice to take the most similar gallery person (denoted as $g$) to the query person $q$ as the candidate for this gallery scene image and rank all candidates from all gallery scene images based on the individual similarity. This common procedure ignores the valuable information of context persons. It is intuitive that if more context persons of the query person $q$ appear in the gallery scene image with him, it is more likely that the $g$ is a right candidate of $q$. However, it is still unknown how to utilize the context persons to rectify the matching degree of the target query-gallery pair $(q,g)$.

In fact, one one hand, when the target query-gallery pair $(q,g)$ is not the same person, any context person should make few contributions to the rectification, even if this context person appears in both the query and gallery scenes. On the other hand, when the target pair $(q,g)$ is the same person, the contributions a context person makes to the rectification should depend on the possibility that this context person also appears in the gallery scene. To model this relationships between the query person $q$ and a context person $q_i$, we propose the co-occurrence index score based on the target query-gallery pair $(q,g)$ and a context query-gallery pair $(q_i,g_j)$. The co-occurrence index score $S_{co}(q,q_i)$ for the query person $q$ and a context person $q_i$ is defined as follows:
\begin{equation}
S_{co}(q,q_i) =\begin{cases}
S_I(q,g)\cdot S_I(q_i,g_j), & S_I(q_i,g_j)\geq b\geq 0 \\
0, & \text{otherwise}
\end{cases}
\label{eq.6}
\end{equation}
where $S_I(q,g)$ is the cosine similarity between person $q$ and $g$, the gallery person $g_j$ is the most similar person to the context person $q_i$ in the gallery scene image, and $b$ is a positive threshold to keep confident context query-gallery pair $(q_i,g_j)$.

Based on the co-occurrence index score, we define the context similarity score $S_C(q,g)$ for the target query-gallery pair $(q,g)$ as follows:
\begin{equation}
    S_C(q,g)=\sum_{i=1}^N S_{co}(q,q_i).
    \label{eq.7}
\end{equation}

During the ranking phase, the context similarity score $S_C(q,g)$ is regarded as an additional cue to rank the gallery persons. Specifically, we combine it with  $S_I(q,g)$ to define the final similarity score $S(q,g)$.
\begin{equation}
    S(q,g)=S_I(q,g)+\lambda \cdot S_C(q,g), 
    \label{eq.8}
\end{equation}
where $\lambda \in (0,1)$ balances the relative importance between $S_I(q,g)$ and $S_C(q,g)$.

\section{Experiments}

In this section, we conduct experiments on two public person search datasets, the PRW~\cite{zheng2017person} and CUHK-SYSU~\cite{xiao2017joint} datasets, and compare the proposed method with state-of-the-art methods. Afterward, thorough ablation studies are conducted to validate the effectiveness of each component. 

\subsection{Datasets}

\textbf{PRW} dataset~\cite{zheng2017person} is collected in a university with six cameras. It provides 11,816 video frames with 43,110 bounding boxes. Among them, 5,704 frames are split
into the training set, 2,057 person bounding boxes are taken as the query set, and 6,112 frames are taken as the gallery set. The training set contains 15,575 bounding boxes from 482 labeled identities and many unlabeled bounding boxes.
Both the training set and gallery sets include plenty of unlabeled bounding boxes which have different identities from those labeled ones. Different from the CUHK-SYSU, the search scope is the whole gallery set, making it more challenging than the CUHK-SYSU.

\textbf{CUHK-SYSU} dataset~\cite{xiao2017joint} contains video frames from the street snap and movies. It provides 18,184 frames with total 96,143 bounding boxes containing both labeled and unlabeled identities. The training set consists of 11,206 frames containing 15,080 bounding boxes from 5,532 labeled identities and many unlabeled bounding boxes, while the testing set is composed of 2,900 query persons and 6,987 gallery frames. For each query person, it provides several gallery subsets with various gallery sizes.

\subsection{Evaluation Protocol}

The widely-used Cumulative Matching Characteristic (CMC) and mean Average Precision (mAP) in the Re-ID task are adopted to evaluate the performance of person search. 
For each query, its AP is scaled by its recall rate. Then the mAP is computed as the average of all APs across all query persons.

\subsection{Implementation Details}

Following the settings proposed in~\cite{luo2019strong}, we pre-train the strong Re-ID teacher model. When training the TDN, the Re-ID teacher model is frozen.
The TDN is trained for total 40 epochs with batch size 16 using the SGD optimizer. The initial learning rate is 0.01, decayed to 0.001 and 0.0001 in the 25-th and 30-th epochs, respectively.
For the RPN in the detection branch, the anchor sizes are set to 4, 8, 16, and 32 for each location on feature maps. Since the height of a person is not less than the width in a scene image, the anchor aspect ratios are empirically set to 1, 2, and 3. The height and width of an input scene image are scaled by the same factor to make the shorter side not less than 640 pixels or the longer side not more than 960 pixels. During reference, the predicted person RoIs with foreground scores lower than 0.5 are removed, and only person RoIs whose Intersection over Union (IoU) with ground truth bounding boxes larger than 0.5 are regarded as true detection results. In addition, the parameters $b$ in (\ref{eq.6}) and $\lambda$ in (\ref{eq.8}) are 0.3 and 0.2, respectively.

\subsection{Comparison with State-of-the-art Methods}

\begin{table}[htp]
\centering
\caption{Comparison with state-of-the-art methods. The gallery size is 100 for the CUHK-SYSU dataset, while the whole gallery set is utilized for the PRW dataset. The strong Re-ID model is used as the teacher model for our method here.}
\resizebox{0.48\textwidth}{!}{%
\begin{tabular}{@{}clcccc@{}}
\toprule
\multicolumn{2}{c}{\multirow{2}{*}{Method}}                & \multicolumn{2}{c}{PRW} & \multicolumn{2}{c}{CUHK-SYSU} \\
\cmidrule(lr){3-4} \cmidrule(l){5-6} 
\multicolumn{2}{c}{}                                       & mAP (\%)  & top-1 (\%)  & mAP (\%)     & top-1 (\%)     \\ \midrule
\multicolumn{1}{c|}{\multirow{5}{*}{\rotatebox{90}{two-step}}}  & DPM+IDE~\cite{zheng2017person}  & 20.5      & 48.3        & -            & -            \\
\multicolumn{1}{c|}{}                                           & MGTS~\cite{chen2018person}      & 32.6      & 72.1        & 83.0         & 83.7         \\
\multicolumn{1}{c|}{}                                           & CLSA~\cite{lan2018person}       & 38.7      & 65.0        & 87.2         & 88.5         \\
\multicolumn{1}{c|}{}                                           & RDLR~\cite{han2019re}           & 42.9      & 70.2        & 93.0         & 94.2         \\
\multicolumn{1}{c|}{}                                           & TCTS~\cite{wang2020tcts}        & 46.8      & 87.5        & 93.9         & 95.1         \\ 
\midrule
\multicolumn{1}{c|}{\multirow{12}{*}{\rotatebox{90}{one-step}}} & OIM~\cite{xiao2017joint}        & -         & -           & 75.7         & 78.7         \\
\multicolumn{1}{c|}{}                                           & IAN~\cite{xiao2019ian}          & 23.0      & 61.9        & 76.3         & 80.1         \\
\multicolumn{1}{c|}{}                                           & NPSM\cite{liu2017neural}        & 24.2      & 53.1        & 77.9         & 81.2         \\
\multicolumn{1}{c|}{}                                           & RCAA~\cite{chang2018rcaa}       & -         & -           & 79.3         & 81.3         \\
\multicolumn{1}{c|}{}                                           & LCGPS~\cite{yan2019learning}    & 33.4      & 73.6        & 84.1         & 86.5         \\
\multicolumn{1}{c|}{}                                           & QEEPS~\cite{munjal2019query}    & 37.1      & 76.7        & 88.9         & 89.1         \\
\multicolumn{1}{c|}{}                                           & NAE+~\cite{chen2020norm}        & 44.0      & 81.1        & 92.1         & 92.9         \\
\multicolumn{1}{c|}{}                                           & APNet~\cite{zhong2020robust}    & 41.9      & 81.4        & 88.9         & 89.3         \\
\multicolumn{1}{c|}{}                                           & IGPN~\cite{dong2020instance}    & 47.2      & 87.0        & 90.3         & 91.4         \\
\multicolumn{1}{c|}{}                                           & BINet~\cite{dong2020bi}         & 45.3      & 81.7        & 90.0         & 90.7         \\
\multicolumn{1}{c|}{}                                           & PSFL~\cite{kim2021prototype}    & 44.2      & 85.2        & 92.3         & 94.7         \\ 
\cmidrule(l){2-6}
\multicolumn{1}{c|}{} & Ours                                    & \bf70.2   & \bf93.5     &  \bf94.9     &  \bf96.3     \\ 
\bottomrule
\end{tabular}%
}
\label{tab:comparison_with_sota}
\end{table}

In this section, we compare the proposed method with some state-of-the-art methods. Experimental results are reported in Table \ref{tab:comparison_with_sota}. In this section, the strong Re-ID model~\cite{luo2019strong} is taken as the default teacher model in our method.

\textbf{Results on PRW.} As shown in Table~\ref{tab:comparison_with_sota}, the proposed method achieves 70.2\% mAP and 93.5\% top-1 recognition rate, outperforming all the compared state-of-the-art methods. Among one-step state-of-the-art methods, the IGPN obtains the highest performance (47.2\% mAP and 87.0\% top-1). Compared to the IGPN, the proposed method achieves 23.0\% mAP and 6.5\% top-1 improvement. We also make comparisons with some two-step methods among which the TCTS method achieves the best performance. According to Table~\ref{tab:comparison_with_sota}, our proposed method outperforms TCTS by 23.4\% in mAP and 6.0\% in top-1 recognition rate respectively. It is observed that the proposed method surpasses all the listed state-of-the-art methods by a large margin, especially in mAP, which shows that the proposed method can retrieve more positive gallery persons at the top of the ranking list. 

\textbf{Results on CUHK-SYSU.} As shown in Table \ref{tab:comparison_with_sota}, the proposed method achieves 94.9\% mAP and 96.3\% top-1 recognition rate, surpassing all the compared state-of-the-art methods. For example, compared to TCTS which achieves the best performance (93.9\% mAP and 95.1\% top-1) among all the compared state-of-the-art methods, the proposed method achieves 1.0\% gains in mAP and 1.2\% gains in top-1 performance. 

\begin{figure}[t]
\begin{center}
   \includegraphics[width=1.0\linewidth]{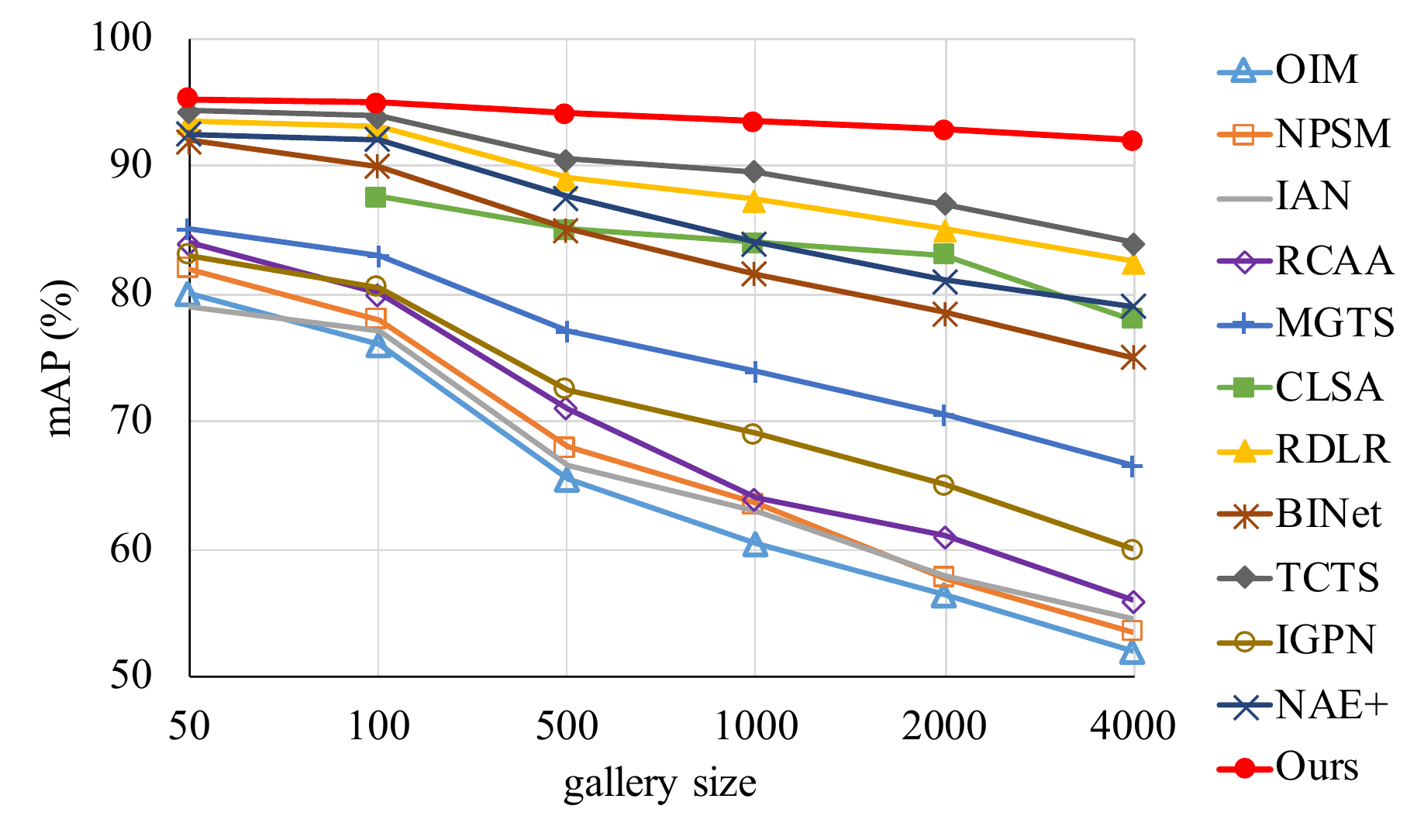}
\end{center}
\caption{Performance comparison on the CUHK-SYSU dataset with gallery size increasing from 50 to 4000.}
\label{fig:map_gallery_sizes}
\end{figure}

Besides, we also conduct experiments to evaluate the influence of various gallery sizes. The gallery sizes vary from 50 to 4,000. Detailed comparison results are presented in Fig.~\ref{fig:map_gallery_sizes}. On one hand, we observe that the performance of all methods decreases with gallery size increasing. On the other hand, the proposed method generates much higher and more stable performance compared with other state-of-the-art methods, especially when the gallery size is large. The robustness against various gallery sizes is valuable in practical scenarios.  

\subsection{Ablation Study}

In this section, experiments are first conducted to validate the effectiveness of each proposed component. Then, the comparison results between the one-step and two-step frameworks are reported. Next, the running speed of the previous one-step framework and the proposed TDN is compared. Finally, the parameter analysis is conducted to study the influence of hyper-parameters in the proposed RCP.
\begin{table}[ht]
\centering
\caption{Effectiveness of the proposed TDN and RCP. ``OIM*'' represents the modified OIM method, the previous widely-used one-step framework.}
\resizebox{0.42\textwidth}{!}{%
\begin{tabular}{@{}lccll@{}}
\toprule
\multirow{2}{*}{Method} & \multicolumn{2}{c}{PRW}                                       & \multicolumn{2}{l}{CUHK-SYSU} \\  
\cmidrule(lr){2-3} \cmidrule(l){4-5} 
                        & \multicolumn{1}{l}{mAP (\%)} & \multicolumn{1}{l}{top-1 (\%)} & mAP (\%)     & top-1 (\%)     \\ \midrule
OIM*                     &  22.8     &  49.1    &  81.5     &  83.2     \\
TDN                      &  68.7     &  91.9    &  93.8     &  94.9    \\
TDN + RCP                &  \bf70.2  &  \bf93.5 &  \bf94.9  &  \bf96.3 \\ \midrule
% \midrule
\end{tabular}%
}
\label{tab:componet analysis}
\end{table}

\textbf{Effectiveness of TDN.} 
To validate the effectiveness of the proposed TDN, we re-implement the widely-used one-step person search framework (denoted as OIM*) proposed in~\cite{xiao2017joint}. For a fair comparison, we replace the original ResNet50 backbone with the ResNet50-IBN-a, and divides the backbone into two parts from the conv3\_4.
As shown in Table~\ref{tab:componet analysis}, on the PRW dataset, the proposed TDN surpasses the OIM* by 45.9\% in mAP and 42.8\% in top-1. On the CUHK-SYSU dataset, the proposed TDN outstrips the OIM* by 12.3\% in mAP and 11.7\% in top-1. This demonstrates that the proposed TDN one-step person search framework can significantly boost the one-step person search performance by integrating the advanced Re-ID model.

In the proposed TDN, to help the one-step person search model learn discriminative Re-ID features from the powerful Re-ID teacher model, we propose to partially disentangle the detection and Re-ID subtasks and introduce the KTB module. 
To validate their effectiveness, we construct a baseline model by removing the teacher branch and KTB module in the proposed TDN. The baseline is also trained end-to-end.
Experimental results are reported in Table~\ref{tab:TDN}. Compared to the OIM*, the proposed baseline achieves better results on both datasets, especially on the more challenging PRW dataset (15.2\% mAP and 28.3\% top-1 improvement).
This demonstrates that the proposed Partially Disentangled Framework can help the one-step person search model learn more discriminative Re-ID features, and can be used as a new strong baseline framework for the one-step person search model in place of the OIM one-step framework. Besides, we also employ the Re-ID teacher model to guide the learning of the OIM* model. As shown in Table~\ref{tab:TDN}, compared to the ``OIM* only w/ teacher", the ``Baseline only w/ teacher" achieves much higher performance on both datasets. For example, on the PRW dataset, the ``Baseline only w/ teacher" surpasses the ``OIM* only w/ teacher" by 12.0\% in mAP and 7.5\% in top-1. This further validates that the proposed Partially Disentangled Framework is effective to help the one-step person search model learn from the powerful Re-ID teacher model.

Without the KTB module, the performance of TDN drops from 68.7\% to 63.0\% in mAP on the PRW dataset and from 93.8\% to 93.1\% in mAP on the CUHK-SYSU dataset. This validates that the KTB module is crucial to help the one-step person search model learn from the teacher model. When the Re-ID teacher branch is removed, the performance of TDN suffers heavy losses. On the PRW dataset, the mAP drops from 68.7\% to 55.2\% and the top-1 drops from 91.9\% to 89.5\%. On the CUHK-SYSU dataset, the mAP drops from 93.8\% to 90.0\% and the top-1 drops from 94.9\% to 91.8\%. This shows that it is effective to improve the one-step person search performance by learning from the powerful Re-ID model.
\begin{table}[t]
\centering
\caption{Impact of components in the proposed TDN.}
\resizebox{0.45\textwidth}{!}{%
\begin{tabular}{@{}lccll@{}}
\toprule
\multirow{2}{*}{Method} & \multicolumn{2}{c}{PRW}                                       & \multicolumn{2}{l}{CUHK-SYSU} \\  
\cmidrule(lr){2-3} \cmidrule(l){4-5} 
                        & \multicolumn{1}{l}{mAP (\%)} & \multicolumn{1}{l}{top-1 (\%)} & mAP (\%)     & top-1 (\%)     \\ \midrule
OIM*                     &  22.8     &  49.1    &  81.5     &  83.2    \\
Baseline                 & \bf38.0   & \bf77.4  & \bf82.4   & \bf84.8    \\ \midrule
OIM* only w/ teacher     &  51.0     &  82.7    &  88.5     &  90.5    \\
Baseline only w/ teacher & \bf63.0   & \bf90.2  & \bf93.1   &  \bf94.5    \\ \midrule
TDN (w/o KTB)            &  63.0     &  90.2    &  93.1     &  94.5    \\
TDN (w/o teacher)        &  55.2     &  89.5    &  90.0     &  91.8    \\
TDN                      &  \bf68.7  &  \bf91.9 &  \bf93.8  &  \bf94.9    \\
\midrule
\end{tabular}%
}
\label{tab:TDN}
\end{table}

\textbf{Effectiveness of RCP.}
As shown in Table~\ref{tab:componet analysis}, by introducing RCP in the ranking phase, the person search performance further witnesses a gain of 1.5\% in mAP and 1.6\% in top-1 recognition rate on the PRW dataset, and 1.1\% in mAP and 1.4\% in top-1 recognition rate on the CUHK-SYSU dataset. It demonstrates that the proposed RCP can effectively improve performance by exploiting context information.

\textbf{Enjoying the merits of person Re-ID.} 
In this paper, the TDN is proposed to make the end-to-end person search enjoy the merits of person Re-ID researches. To demonstrate that the proposed TDN is able to realize this purpose, besides employing the strong Re-ID baseline model proposed by~\cite{luo2019strong} as the teacher model, we also adopt some other state-of-the-art Re-ID models as the teacher model, the MGN~\cite{wang2018learning} model and AGW model~\cite{ye2021deep}. Experimental results are reported in Table~\ref{tab:more_reid_teacher_model}. It is observed that the proposed TDN achieves much higher performance on both person search datasets with any one of the three state-of-the-art Re-ID models as the teacher model when compared to the baseline model.

The MGN model is designed to extract multiple granularity locals features as well as a global feature for a person patch. The final feature of a person patch is obtained by concatenating the global feature and multiple granularity global features. Although the TDN has no local branches to learn local features for person RoIs, it can still learn discriminative Re-ID features with the guidance of the MGN teacher model.
The AGW model integrates the Non-local Attention Block into the backbone networks to capture non-local relations and replaces the widely-used max-pooling or average pooling with a learnable generalized-mean pooling layer to capture the fine-grained discriminative features. Even if the TDN has no Non-local Attention Blocks and the generalized-mean pooling layer to learn the non-local relations between pixels and fine-grained features, it can also learn discriminative Re-ID features from the AGW teacher model. These experimental results demonstrate that the proposed TDN can significantly boost the performance of the end-to-end person search by enjoying the merits of state-of-the-art person Re-ID models.

\begin{table}[t]
\centering
\caption{Results of different state-of-the-art Re-ID models as the teacher model in the proposed TDN framework. The ``None" means the baseline model where no teacher model is applied. The proposed RCP ranking method is not used here.}
\resizebox{0.45\textwidth}{!}{%
\begin{tabular}{@{}lccll@{}}
\toprule
\multirow{2}{*}{Teacher model} & \multicolumn{2}{c}{PRW}                                       & \multicolumn{2}{l}{CUHK-SYSU} \\  
\cmidrule(lr){2-3} \cmidrule(l){4-5} 
                        & \multicolumn{1}{l}{mAP (\%)} & \multicolumn{1}{l}{top-1 (\%)} & mAP (\%)     & top-1 (\%)     \\ \midrule
None (baseline)                  &  38.0     &  77.4    &  82.4     &  84.8    \\  
Strong Re-ID~\cite{luo2019strong}   &  68.7     &  91.9    &  93.8     &  94.9    \\
MGN~\cite{wang2018learning}      &  71.4     &  92.7    &  93.5     &  95.1    \\
AGW~\cite{ye2021deep}            &  68.9     &  93.1    &  92.9     &  94.5    \\ \midrule
\end{tabular}%
}
\label{tab:more_reid_teacher_model}
\end{table}

\textbf{One-step vs. two-step.}
A straightforward solution to person search is the two-step methods which employ a person detector to detect person RoIs and use a person Re-ID model to extract feature representations for the predicted person RoIs. To validate that the proposed one-step person search model TDN is a better solution to person search task compared to the two-step ones, we construct several two-step methods by training a Faster R-CNN person detector and employing different Re-ID teacher model to extract features for the predicted person RoIs. 
Comparison results are reported in Table~\ref{tab:one-step vs. two-step}.  All the three two-step methods achieve pretty high performance on both datasets with the advanced Re-ID models as the feature extractor. It is also observed that these two-step methods outperforms most state-of-the-art person search methods, which shows that the state-of-the-art Re-ID models are very powerful when applied to person search task.
However, when adopting any one of these advanced Re-ID models as the teacher model, the proposed TDN realizes higher performance than the corresponding two-step method. It validates that the proposed TDN can not only learn from the powerful Re-ID teacher models but also can be more excellent than the teacher by exploiting the correlations between two subtasks in the end-to-end one-step framework.
\begin{table}[t]
\centering
\caption{Comparison of one-step and two-step methods. The ``TDN (MGN)" represents the TDN model with the MGN Re-ID model as the teacher. The proposed RCP ranking method is not used here.}
\resizebox{0.48\textwidth}{!}{%
\begin{tabular}{@{}clcccc@{}}
\toprule
\multicolumn{2}{c}{\multirow{2}{*}{Method}}                & \multicolumn{2}{c}{PRW} & \multicolumn{2}{c}{CUHK-SYSU} \\
\cmidrule(lr){3-4} \cmidrule(l){5-6} 
\multicolumn{2}{c}{}           & mAP (\%)  & top-1 (\%)  & mAP (\%)     & top-1 (\%)     \\ \midrule
& Faster R-CNN + Strong Re-ID  & 66.9      & 91.3        & 92.1         & 94.1    \\
& TDN (Strong Re-ID)           & \bf68.7   & \bf91.9     & \bf93.8      & \bf94.9    \\\midrule
& Faster R-CNN + MGN           & 66.8      & 92.3        & 90.5         & 92.6    \\
& TDN (MGN)                    & \bf71.4   & \bf92.7     & \bf93.5      & \bf95.1    \\\midrule
& Faster R-CNN + AGW           & 67.1      & 91.3        & 91.4         & 93.4    \\
& TDN (AGW)                    & \bf68.9   & \bf93.1     & \bf92.9      & \bf94.5    \\ \bottomrule
\end{tabular}%
}
\label{tab:one-step vs. two-step}
\end{table}

\textbf{Running Speed.} Most of the previous one-step methods adopt the Faster R-CNN-based one-step framework (denoted as OIM*) and design more additional modules based on it. This means that the running speed of these methods is slower than the OIM*. Thus, in Table~\ref{tab:running-speed}, we only compare the running speed of the OIM* and the proposed TDN. It is observed that the proposed TDN runs three times as fast as the OIM*. In the OIM* framework, the second part has to process the RoI feature maps from all the RoIs generated by the RPN (e.g. 128 RoIs per image). However, the RoIs generated by the RPN contain a lot of invalid proposals which do not contain persons or are removed after the NMS. The computation on invalid RoIs greatly lowers the running speed. In contrast, in the proposed TDN, the Part-2 in the Re-ID branch only processes the final detections generated by the lightweight detection branch for each scene image. Assuming 10 final predicted person bounding boxes in each frame on average, we quantitatively compare the computation complexity of main parts for two one-step frameworks. As shown in Table~\ref{tab:compution}, the computation of the proposed TDN is much lower than the previous one-step framework OIM*. Thus, the proposed TDN can run much faster than the OIM*.

\begin{table}[htp]
\centering
\caption{Running speed comparison of the OIM* and TDN.}
\resizebox{0.36\textwidth}{!}{%
\begin{tabular}{@{}lccc@{}}
\toprule
Method & GPU         & PRW  & CUHK-SYSU \\ \midrule
OIM*   & TITAN RTX   & 11.5 fps         & 12.4 fps               \\
TDN    & TITAN RTX   & 34.3 fps         & 37.7 fps               \\ \bottomrule
\end{tabular}%
}
\label{tab:running-speed}
\end{table}

\begin{table}[t]
\centering
\caption{Computation complexity comparison of two one-step frameworks. The numbers \textcolor{red}{128} and \textcolor{red}{10} are the numbers of RoIs from the RPN and assumed final predicted person bounding boxes, respectively.}
\resizebox{0.48\textwidth}{!}{%
\begin{tabular}{@{}lcccccc@{}}
\toprule
\begin{tabular}[c]{@{}c@{}}Computation \\ (GFLOPs)\end{tabular} & Part-1         & RoI Head       & KTB         & Part-2          & Total  \\ \midrule
input size          & 960x600x3  & 7x7x512        & 256x128x3   & 32x16x512       &        \\ \midrule
OIM*        & 42.2             & -              & -           & \textcolor{red}{128}x2.9=371.2 & 413.4 \\
TDN        & 42.2          & \textcolor{red}{128}x0.05=6.4 & \textcolor{red}{10}x2.4=24.0 & \textcolor{red}{10}x2.9=29.0     & 101.6 \\ \bottomrule
\end{tabular}%
}
\label{tab:compution}
\end{table}

\textbf{Parameter Analysis.} We conduct experiments to evaluate the impact of parameters $b$ in (\ref{eq.5}) and $\lambda$ in (\ref{eq.6}).

\begin{figure}[htp]
\begin{center}
   \includegraphics[width=1.0\linewidth]{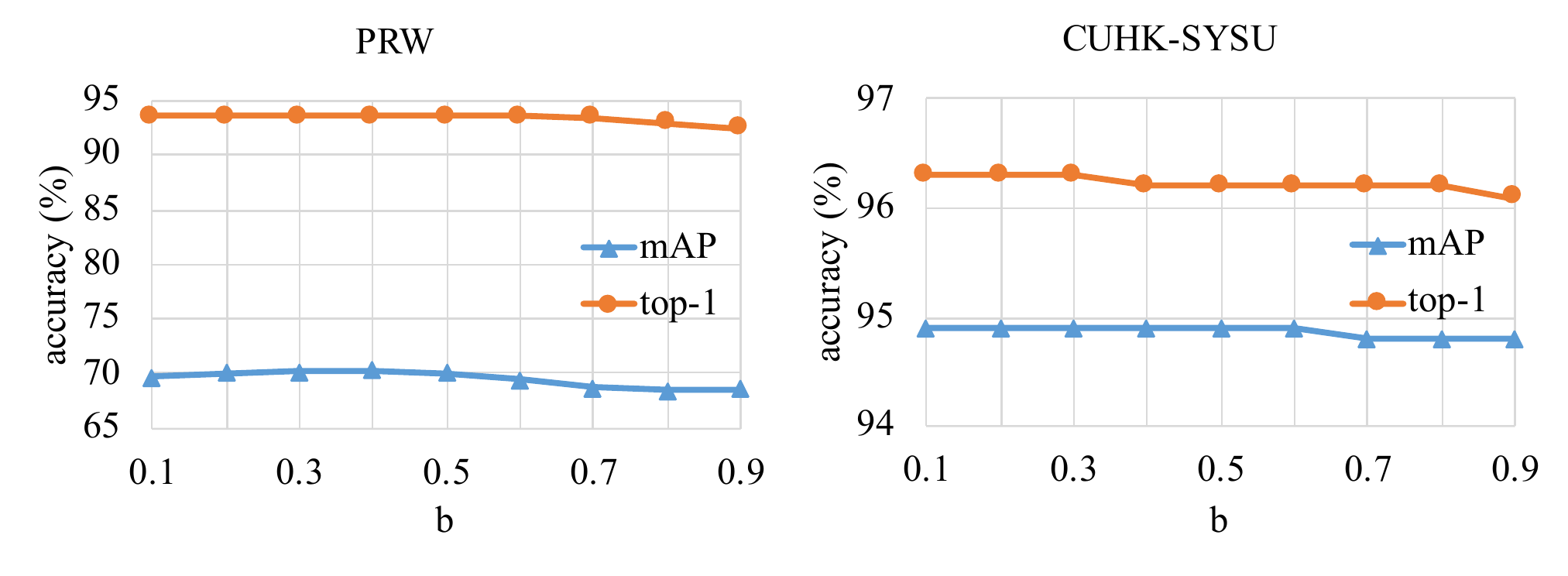}
\end{center}
%\vspace{-0.3cm}
\caption{Impact of parameter $b$. $\lambda$ is set to 0.2 for both datasets.}
\label{fig:impact of b}
\end{figure}

As shown in Fig.~\ref{fig:impact of b}, on the PRW dataset, both the mAP and top-1 can obtain stable gains when the threshold $b\leq 0.6$, and when $b>0.6$, the mAP and top-1 begins to decrease slightly. On the CUHK-SYSU dataset, when $b\leq 0.3$, the performance achieves the highest improvement, and when $b\geq 0.3$, the improvement becomes smaller.
Generally, the proposed RCP ranking method is not sensitive to the threshold $b$. In a large reasonable range, any value of $b$ can bring performance gains. It is recommended to set $b$ to 0.3, considering the performance on both datasets.

\begin{figure}[htp]
\begin{center}
   \includegraphics[width=1.0\linewidth]{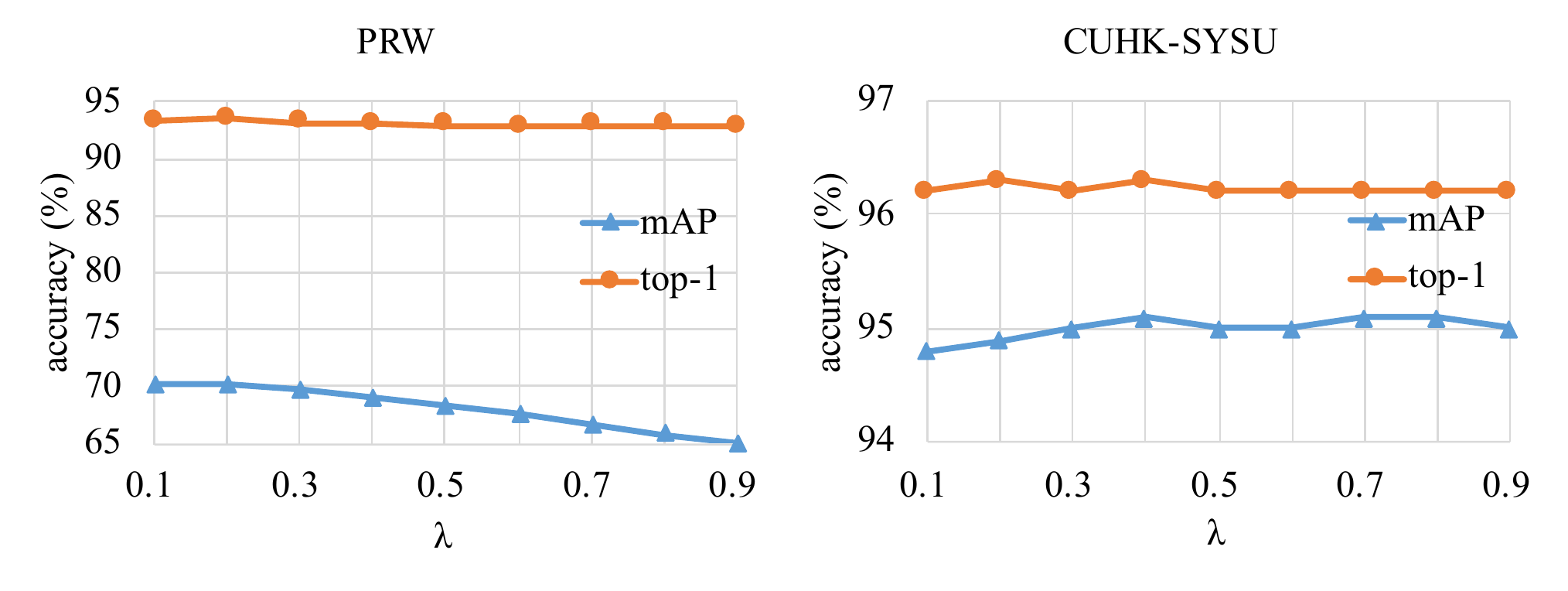}
\end{center}
\caption{Impact of parameter $\lambda$. $b$ is set to 0.3 for both datasets.}
\label{fig:impact of lambda}
\end{figure}

Fig.~\ref{fig:impact of lambda} shows the impact of $\lambda$. On the PRW dataset, when $\lambda \in [0.1, 0.3]$, the proposed RCP brings performance improvement. On the CUHK-SYSU dataset, any value of $\lambda \in [0.1, 0.9]$ can improve the performance. Experimental results show that the proposed RCP is not sensitive to $\lambda \in [0.1, 0.3]$ on two datasets. In this paper, we set $\lambda$ to 0.2 for both datasets.

\subsection{Visualization}
\begin{figure*}[htp]
\begin{center}
   \includegraphics[width=1.0\linewidth]{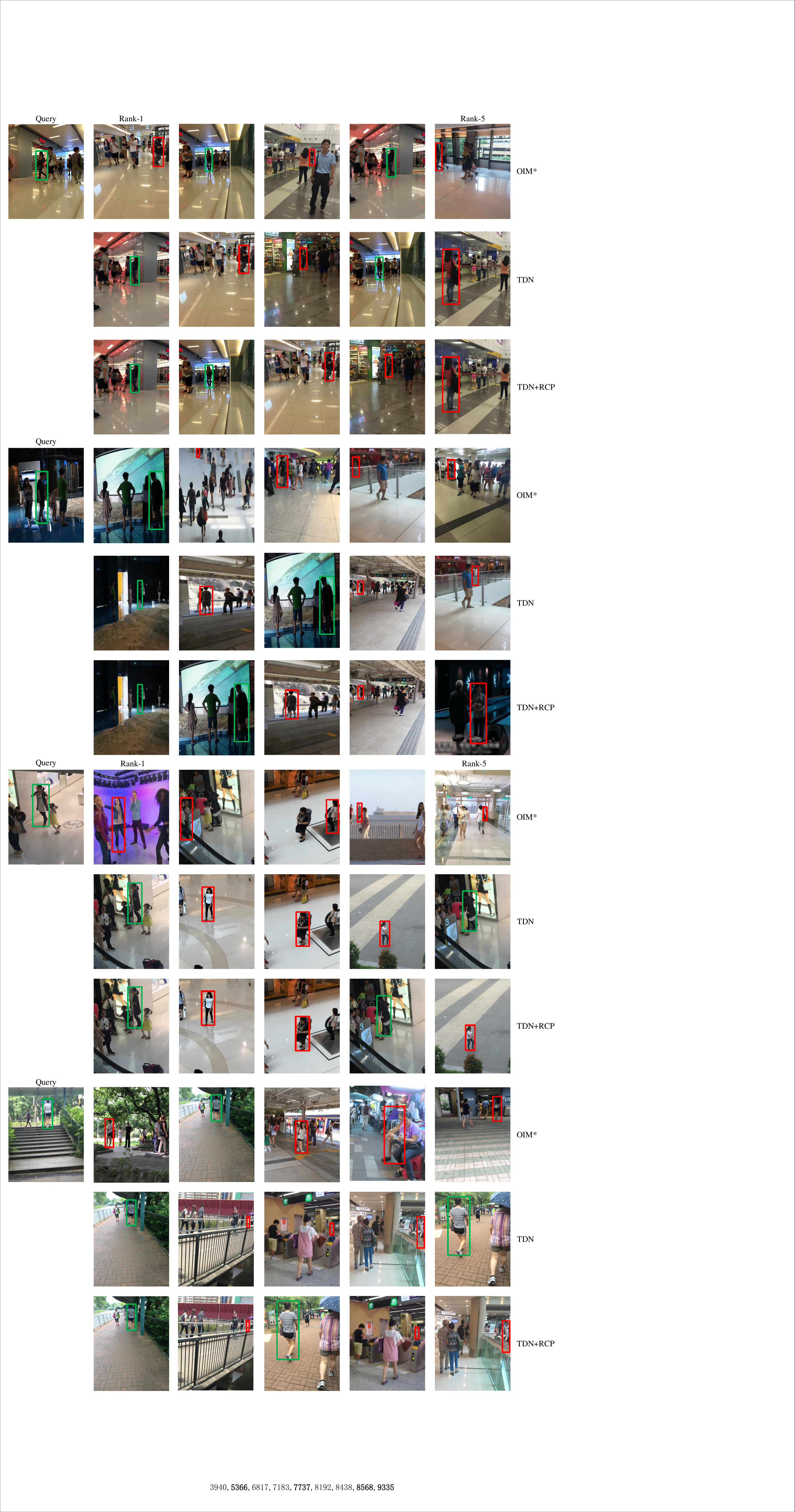}
\end{center}
%\vspace{-0.3cm}
\caption{Example results of two query persons. The query persons are marked with green bounding boxes in the first column scene images. For each query person, the first row, the second row and the third row ranking results correspond to the baseline method OIM*, the TDN method and the ``TDN + RCP" method, respectively. In each ranking scene image, the most similar gallery person to the query person is marked within a bounding box. The green bounding box is for the right match of the query person, and the red one is for the wrong match.}
\label{fig:visualization1}
\end{figure*}

\begin{figure*}[htp]
\begin{center}
   \includegraphics[width=1.0\linewidth]{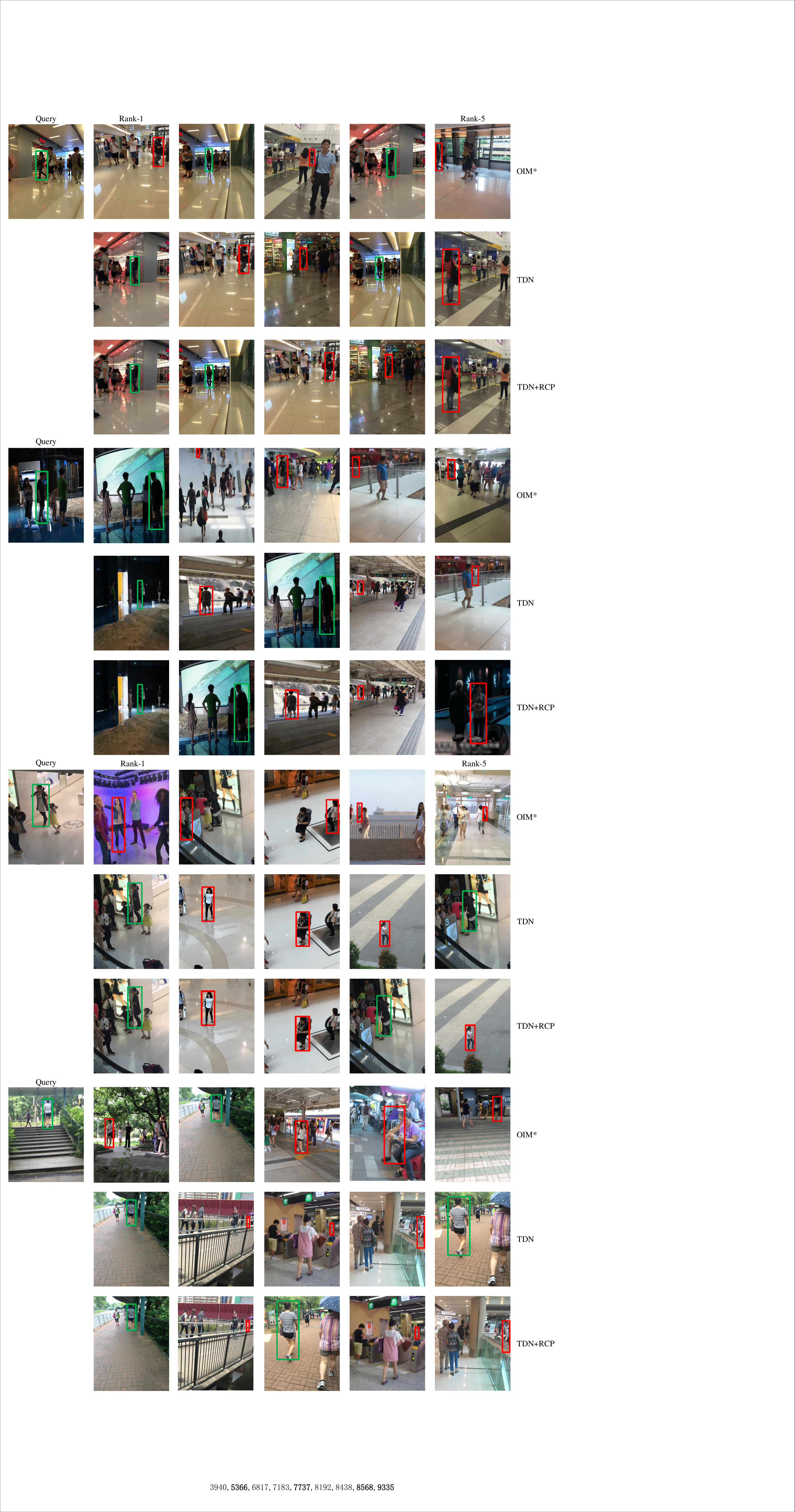}
\end{center}
%\vspace{-0.3cm}
\caption{Example results of another two query persons. The query persons are marked with green bounding boxes in the first column scene images. For each query person, the first row, the second row and the third row ranking results correspond to the baseline method OIM*, the TDN method and the ``TDN + RCP" method, respectively. In each ranking scene image, the most similar gallery person to the query person is marked within a bounding box. The green bounding box is for the right match of the query person, and the red one is for the wrong match.}
\label{fig:visualization2}
\end{figure*}

To further validate the effectiveness of the proposed TDN model as well as the RCP ranking method, some visualization results are reported in Fig.~\ref{fig:visualization1} and Fig.~\ref{fig:visualization2}. Compared to the previous widely-used OIM*, the proposed TDN can retrieve more true candidates at the top of the ranking list. When the proposed RCP ranking method is applied, the quality of the ranking list is further improved. These visualization results demonstrate the effectiveness of the proposed TDN and RCP.

\section{Conclusion}

In this paper, aiming to make the one-step person search model enjoy the merits of the state-of-the-art Re-ID models, we propose the Teacher-guided Disentangling Networks which is a faster and stronger one-step person search framework. The proposed TDN presents a pipeline to integrate the prior knowledge from the powerful Re-ID model to boost the person search performance. In the proposed TDN, to make the one-step person search model learn discriminative Re-ID features from the powerful Re-ID teacher model better, we propose to partially disentangle the Partially Disentangled Framework and introduce the Knowledge Transfer Bridge module. Besides, we propose the Ranking with Context Persons strategy to exploit the context persons provided in person search task. The proposed RCP ranking method can further improve the person search performance. Comparison results with the previous state-of-the-art methods, as well as thorough ablation studies, demonstrate the favorable performance of the proposed method.

% if have a single appendix:
%\appendix[Proof of the Zonklar Equations]
% or
%\appendix  % for no appendix heading
% do not use \section anymore after \appendix, only \section*
% is possibly needed

% use appendices with more than one appendix
% then use \section to start each appendix
% you must declare a \section before using any
% \subsection or using \label (\appendices by itself
% starts a section numbered zero.)
%

% use section* for acknowledgment

% Can use something like this to put references on a page
% by themselves when using endfloat and the captionsoff option.
\ifCLASSOPTIONcaptionsoff
  \newpage
\fi

% trigger a \newpage just before the given reference
% number - used to balance the columns on the last page
% adjust value as needed - may need to be readjusted if
% the document is modified later
%\IEEEtriggeratref{8}
% The "triggered" command can be changed if desired:
%\IEEEtriggercmd{\enlargethispage{-5in}}

% references section

% can use a bibliography generated by BibTeX as a .bbl file
% BibTeX documentation can be easily obtained at:
% http://mirror.ctan.org/biblio/bibtex/contrib/doc/
% The IEEEtran BibTeX style support page is at:
% http://www.michaelshell.org/tex/ieeetran/bibtex/
%\bibliographystyle{IEEEtran}
% argument is your BibTeX string definitions and bibliography database(s)
%\bibliography{IEEEabrv,../bib/paper}

\begin{thebibliography}{10}
\providecommand{\url}[1]{#1}
\csname url@samestyle\endcsname
\providecommand{\newblock}{\relax}
\providecommand{\bibinfo}[2]{#2}
\providecommand{\BIBentrySTDinterwordspacing}{\spaceskip=0pt\relax}
\providecommand{\BIBentryALTinterwordstretchfactor}{4}
\providecommand{\BIBentryALTinterwordspacing}{\spaceskip=\fontdimen2\font plus
\BIBentryALTinterwordstretchfactor\fontdimen3\font minus
  \fontdimen4\font\relax}
\providecommand{\BIBforeignlanguage}[2]{{%
\expandafter\ifx\csname l@#1\endcsname\relax
\typeout{** WARNING: IEEEtran.bst: No hyphenation pattern has been}%
\typeout{** loaded for the language `#1'. Using the pattern for}%
\typeout{** the default language instead.}%
\else
\language=\csname l@#1\endcsname
\fi
#2}}
\providecommand{\BIBdecl}{\relax}
\BIBdecl

\bibitem{hinton2015distilling}
G.~Hinton, O.~Vinyals, and J.~Dean, ``Distilling the knowledge in a neural
  network,'' \emph{arXiv preprint arXiv:1503.02531}, 2015.

\bibitem{chen2018person}
D.~Chen, S.~Zhang, W.~Ouyang, J.~Yang, and Y.~Tai, ``Person search via a
  mask-guided two-stream cnn model,'' in \emph{Proceedings of the European
  Conference on Computer Vision}, 2018, pp. 734--750.

\bibitem{goodfellow2016deep}
I.~Goodfellow, Y.~Bengio, A.~Courville, and Y.~Bengio, \emph{Deep
  learning}.\hskip 1em plus 0.5em minus 0.4em\relax MIT press Cambridge, 2016,
  vol.~1.

\bibitem{bak2017one}
S.~Bak and P.~Carr, ``One-shot metric learning for person re-identification,''
  in \emph{Proceedings of the IEEE Conference on Computer Vision and Pattern
  Recognition}, 2017, pp. 2990--2999.

\bibitem{liu2017stepwise}
Z.~Liu, D.~Wang, and H.~Lu, ``Stepwise metric promotion for unsupervised video
  person re-identification,'' in \emph{Proceedings of the IEEE International
  Conference on Computer Vision}, 2017, pp. 2429--2438.

\bibitem{hermans2017defense}
A.~Hermans, L.~Beyer, and B.~Leibe, ``In defense of the triplet loss for person
  re-identification,'' \emph{arXiv preprint arXiv:1703.07737}, 2017.

\bibitem{chen2017beyond}
W.~Chen, X.~Chen, J.~Zhang, and K.~Huang, ``Beyond triplet loss: a deep
  quadruplet network for person re-identification,'' in \emph{Proceedings of
  the IEEE Conference on Computer Vision and Pattern Recognition}, 2017, pp.
  403--412.

\bibitem{yu2019reference}
M.~Yu, Z.~Chang, Q.~Zhou, S.~Zheng, and T.~P. Wu, ``Reference-oriented loss for
  person re-identification,'' in \emph{2019 International Joint Conference on
  Neural Networks (IJCNN)}.\hskip 1em plus 0.5em minus 0.4em\relax IEEE, 2019,
  pp. 1--8.

\bibitem{chang2019distribution}
Z.~Chang, Q.~Zhou, M.~Yu, S.~Zheng, H.~Yang, and T.-P. Wu, ``Distribution
  context aware loss for person re-identification,'' \emph{arXiv preprint
  arXiv:1911.07273}, 2019.

\bibitem{yan2021beyond}
C.~Yan, G.~Pang, X.~Bai, C.~Liu, N.~Xin, L.~Gu, and J.~Zhou, ``Beyond triplet
  loss: person re-identification with fine-grained difference-aware pairwise
  loss,'' \emph{IEEE Transactions on Multimedia}, 2021.

\bibitem{chen2018group}
D.~Chen, D.~Xu, H.~Li, N.~Sebe, and X.~Wang, ``Group consistent similarity
  learning via deep crf for person re-identification,'' in \emph{Proceedings of
  the IEEE Conference on Computer Vision and Pattern Recognition}, 2018, pp.
  8649--8658.

\bibitem{chen2018video}
D.~Chen, H.~Li, T.~Xiao, S.~Yi, and X.~Wang, ``Video person re-identification
  with competitive snippet-similarity aggregation and co-attentive snippet
  embedding,'' in \emph{Proceedings of the IEEE Conference on Computer Vision
  and Pattern Recognition}, 2018, pp. 1169--1178.

\bibitem{song2018region}
G.~Song, B.~Leng, Y.~Liu, C.~Hetang, and S.~Cai, ``Region-based quality
  estimation network for large-scale person re-identification,'' in
  \emph{Proceedings of the AAAI Conference on Artificial Intelligence},
  vol.~32, no.~1, 2018.

\bibitem{chang2020weighted}
Z.~Chang, Z.~Qin, H.~Fan, H.~Su, H.~Yang, S.~Zheng, and H.~Ling, ``Weighted
  bilinear coding over salient body parts for person re-identification,''
  \emph{Neurocomputing}, vol. 407, pp. 454--464, 2020.

\bibitem{luo2019strong}
H.~Luo, W.~Jiang, Y.~Gu, F.~Liu, X.~Liao, S.~Lai, and J.~Gu, ``A strong
  baseline and batch normalization neck for deep person re-identification,''
  \emph{IEEE Transactions on Multimedia}, vol.~22, no.~10, pp. 2597--2609,
  2019.

\bibitem{chen2020salience}
X.~Chen, C.~Fu, Y.~Zhao, F.~Zheng, J.~Song, R.~Ji, and Y.~Yang,
  ``Salience-guided cascaded suppression network for person
  re-identification,'' in \emph{Proceedings of the IEEE/CVF Conference on
  Computer Vision and Pattern Recognition}, 2020, pp. 3300--3310.

\bibitem{zhang2020relation}
Z.~Zhang, C.~Lan, W.~Zeng, X.~Jin, and Z.~Chen, ``Relation-aware global
  attention for person re-identification,'' in \emph{Proceedings of the
  IEEE/CVF Conference on Computer Vision and Pattern Recognition}, 2020, pp.
  3186--3195.

\bibitem{gong2021lag}
X.~Gong, Z.~Yao, X.~Li, Y.~Fan, B.~Luo, J.~Fan, and B.~Lao, ``Lag-net:
  Multi-granularity network for person re-identification via local attention
  system,'' \emph{IEEE Transactions on Multimedia}, 2021.

\bibitem{zheng2017person}
L.~Zheng, H.~Zhang, S.~Sun, M.~Chandraker, Y.~Yang, and Q.~Tian, ``Person
  re-identification in the wild,'' in \emph{Proceedings of the IEEE Conference
  on Computer Vision and Pattern Recognition}, 2017, pp. 1367--1376.

\bibitem{lan2018person}
X.~Lan, X.~Zhu, and S.~Gong, ``Person search by multi-scale matching,'' in
  \emph{Proceedings of the European Conference on Computer Vision}, 2018, pp.
  536--552.

\bibitem{han2019re}
C.~Han, J.~Ye, Y.~Zhong, X.~Tan, C.~Zhang, C.~Gao, and N.~Sang, ``Re-id driven
  localization refinement for person search,'' in \emph{Proceedings of the IEEE
  International Conference on Computer Vision}, 2019, pp. 9814--9823.

\bibitem{wang2020tcts}
C.~Wang, B.~Ma, H.~Chang, S.~Shan, and X.~Chen, ``Tcts: A task-consistent
  two-stage framework for person search,'' in \emph{Proceedings of the IEEE/CVF
  Conference on Computer Vision and Pattern Recognition}, 2020, pp.
  11\,952--11\,961.

\bibitem{ren2015faster}
S.~Ren, K.~He, R.~Girshick, and J.~Sun, ``Faster r-cnn: Towards real-time
  object detection with region proposal networks,'' in \emph{Advances in neural
  information processing systems}, 2015, pp. 91--99.

\bibitem{xiao2017joint}
T.~Xiao, S.~Li, B.~Wang, L.~Lin, and X.~Wang, ``Joint detection and
  identification feature learning for person search,'' in \emph{Proceedings of
  the IEEE Conference on Computer Vision and Pattern Recognition}, 2017, pp.
  3415--3424.

\bibitem{li2018spatial}
L.~Li, H.~Yang, and L.~Chen, ``Spatial invariant person search network,'' in
  \emph{Chinese Conference on Pattern Recognition and Computer Vision
  (PRCV)}.\hskip 1em plus 0.5em minus 0.4em\relax Springer, 2018, pp. 122--133.

\bibitem{munjal2019query}
B.~Munjal, S.~Amin, F.~Tombari, and F.~Galasso, ``Query-guided end-to-end
  person search,'' in \emph{Proceedings of the IEEE Conference on Computer
  Vision and Pattern Recognition}, 2019, pp. 811--820.

\bibitem{hong2019cascaded}
Y.~Hong, H.~Yang, L.~Li, L.~Chen, and C.~Liu, ``A cascaded multitask network
  with deformable spatial transform on person search,'' \emph{International
  Journal of Advanced Robotic Systems}, vol.~16, no.~3, p. 1729881419858162,
  2019.

\bibitem{dong2020bi}
W.~Dong, Z.~Zhang, C.~Song, and T.~Tan, ``Bi-directional interaction network
  for person search,'' in \emph{Proceedings of the IEEE/CVF Conference on
  Computer Vision and Pattern Recognition}, 2020, pp. 2839--2848.

\bibitem{chen2020norm}
D.~Chen, S.~Zhang, J.~Yang, and B.~Schiele, ``Norm-aware embedding for
  efficient person search,'' in \emph{Proceedings of the IEEE/CVF Conference on
  Computer Vision and Pattern Recognition}, 2020, pp. 12\,615--12\,624.

\bibitem{romero2014fitnets}
A.~Romero, N.~Ballas, S.~E. Kahou, A.~Chassang, C.~Gatta, and Y.~Bengio,
  ``Fitnets: Hints for thin deep nets,'' \emph{arXiv preprint arXiv:1412.6550},
  2014.

\bibitem{zagoruyko2016paying}
S.~Zagoruyko and N.~Komodakis, ``Paying more attention to attention: Improving
  the performance of convolutional neural networks via attention transfer,''
  \emph{arXiv preprint arXiv:1612.03928}, 2016.

\bibitem{yim2017gift}
J.~Yim, D.~Joo, J.~Bae, and J.~Kim, ``A gift from knowledge distillation: Fast
  optimization, network minimization and transfer learning,'' in
  \emph{Proceedings of the IEEE Conference on Computer Vision and Pattern
  Recognition}, 2017, pp. 4133--4141.

\bibitem{munjal2019knowledge}
B.~Munjal, F.~Galasso, and S.~Amin, ``Knowledge distillation for end-to-end
  person search,'' \emph{arXiv preprint arXiv:1909.01058}, 2019.

\bibitem{pan2018two}
X.~Pan, P.~Luo, J.~Shi, and X.~Tang, ``Two at once: Enhancing learning and
  generalization capacities via ibn-net,'' in \emph{Proceedings of the European
  Conference on Computer Vision}, 2018, pp. 464--479.

\bibitem{he2017mask}
K.~He, G.~Gkioxari, P.~Doll{\'a}r, and R.~Girshick, ``Mask r-cnn,'' in
  \emph{Proceedings of the IEEE international conference on computer vision},
  2017, pp. 2961--2969.

\bibitem{mazzon2013detection}
R.~Mazzon, F.~Poiesi, and A.~Cavallaro, ``Detection and tracking of groups in
  crowd,'' in \emph{2013 10th IEEE International Conference on Advanced Video
  and Signal Based Surveillance}.\hskip 1em plus 0.5em minus 0.4em\relax IEEE,
  2013, pp. 202--207.

\bibitem{xiao2019ian}
J.~Xiao, Y.~Xie, T.~Tillo, K.~Huang, Y.~Wei, and J.~Feng, ``Ian: the individual
  aggregation network for person search,'' \emph{Pattern Recognition}, vol.~87,
  pp. 332--340, 2019.

\bibitem{liu2017neural}
H.~Liu, J.~Feng, Z.~Jie, K.~Jayashree, B.~Zhao, M.~Qi, J.~Jiang, and S.~Yan,
  ``Neural person search machines,'' in \emph{Proceedings of the IEEE
  International Conference on Computer Vision}, 2017, pp. 493--501.

\bibitem{chang2018rcaa}
X.~Chang, P.-Y. Huang, Y.-D. Shen, X.~Liang, Y.~Yang, and A.~G. Hauptmann,
  ``Rcaa: Relational context-aware agents for person search,'' in
  \emph{Proceedings of the European Conference on Computer Vision}, 2018, pp.
  84--100.

\bibitem{yan2019learning}
Y.~Yan, Q.~Zhang, B.~Ni, W.~Zhang, M.~Xu, and X.~Yang, ``Learning context graph
  for person search,'' in \emph{Proceedings of the IEEE Conference on Computer
  Vision and Pattern Recognition}, 2019, pp. 2158--2167.

\bibitem{zhong2020robust}
Y.~Zhong, X.~Wang, and S.~Zhang, ``Robust partial matching for person search in
  the wild,'' in \emph{Proceedings of the IEEE/CVF Conference on Computer
  Vision and Pattern Recognition}, 2020, pp. 6827--6835.

\bibitem{dong2020instance}
W.~Dong, Z.~Zhang, C.~Song, and T.~Tan, ``Instance guided proposal network for
  person search,'' in \emph{Proceedings of the IEEE/CVF Conference on Computer
  Vision and Pattern Recognition}, 2020, pp. 2585--2594.

\bibitem{kim2021prototype}
H.~Kim, S.~Joung, I.-J. Kim, and K.~Sohn, ``Prototype-guided saliency feature
  learning for person search,'' in \emph{Proceedings of the IEEE/CVF Conference
  on Computer Vision and Pattern Recognition}, 2021, pp. 4865--4874.

\bibitem{wang2018learning}
G.~Wang, Y.~Yuan, X.~Chen, J.~Li, and X.~Zhou, ``Learning discriminative
  features with multiple granularities for person re-identification,'' in
  \emph{Proceedings of the 26th ACM international conference on Multimedia},
  2018, pp. 274--282.

\bibitem{ye2021deep}
M.~Ye, J.~Shen, G.~Lin, T.~Xiang, L.~Shao, and S.~C. Hoi, ``Deep learning for
  person re-identification: A survey and outlook,'' \emph{IEEE Transactions on
  Pattern Analysis and Machine Intelligence}, 2021.

\end{thebibliography}
%
% <OR> manually copy in the resultant .bbl file
% set second argument of \begin to the number of references
% (used to reserve space for the reference number labels box)
% Generated by IEEEtran.bst, version: 1.12 (2007/01/11)

\bibliographystyle{IEEEtran}

% biography section
% 
% If you have an EPS/PDF photo (graphicx package needed) extra braces are
% needed around the contents of the optional argument to biography to prevent
% the LaTeX parser from getting confused when it sees the complicated
% \includegraphics command within an optional argument. (You could create
% your own custom macro containing the \includegraphics command to make things
% simpler here.)
%\begin{IEEEbiography}[{\includegraphics[width=1in,height=1.25in,clip,keepaspectratio]{mshell}}]{Michael Shell}
% or if you just want to reserve a space for a photo:

% % insert where needed to balance the two columns on the last page with
% % biographies
% %\newpage

% \begin{IEEEbiographynophoto}{Jane Doe}
% Biography text here.
% \end{IEEEbiographynophoto}

% You can push biographies down or up by placing
% a \vfill before or after them. The appropriate
% use of \vfill depends on what kind of text is
% on the last page and whether or not the columns
% are being equalized.

%\vfill

% Can be used to pull up biographies so that the bottom of the last one
% is flush with the other column.
%\enlargethispage{-5in}

% that's all folks
\end{document}